# Entropy-based coordination for maintenance tasks of an autonomous mobile robot system *

Tapio Heikkilä, *Senior Member, IEEE* , Eric Halbach, Jukka Koskinen and Janne Saukkoriipi,
VTT Technical Research Centre of Finland Ltd.

*Abstract*— **Maintenance tasks represent a potential area for applying multi-purpose Autonomous Mobile Robots (AMRs). Intelligent control and coordination of such a system is challenging and optimization methods are feasible only for small fleets. Decentralized control can provide flexibility and robustness, which are better applicable also for large fleets, though with less guaranteed performance. Our focus is on flexibility and robustness in task scheduling and task assignments and we use entropy as an indirect performance criterion for coordination, both at the system level (maximize entropy) and at the AMR level (minimize entropy). As a distributed coordination scheme, we use a modified contract negotiation protocol. We show preliminarily the feasibility of our approach with simulation results.**

## I. Introduction

Multi-robot systems offer potential benefits such as greater spatial coverage and task versatility compared with systems consisting of a single robot or agent. These systems can consist of simple robots with only a single or small number of functions, or multi-purpose robots which can cooperate to execute complex tasks. In logistics settings, increased autonomy in navigation has resulted in the widespread use of Autonomous Mobile Robots (AMR) for material handling tasks [1]. A robot is considered to be autonomous when it has a perception system providing it with sufficient information about its current state and the environment, and a cognition system that allows it to make its own decisions based on the perception information [2].

A good coordination strategy is essential to make the best use of a multi-robot system [3]. These can be divided into explicit and implicit strategies. In explicit coordination strategies, direct communication occurs between all agents, either via unicast or broadcast messaging. Implicit methods rely on the individual robot's ability to explore, gather and process sensor information, model the environment and reason about the behavior of other agents [3]. The robot then independently makes decisions based on the information is has gathered and processed.

Multi-robot coordination generally has two approaches. One is a centralized strategy in which a single robot or agent processes information from and sends commands to all robots in the system. The other is a distributed approach where each robot does its own planning [4]. Centralized approaches can in principle create globally optimal solutions, however they suffer from drawbacks such as intractability for large systems, communication problems, and a single point of failure. Distributed systems overcome these problems, but can only produce suboptimal solutions. A popular implementation for distributed systems is the market-based approach whereby self-interested agents or groups implement an opportunistic form of efficiency that can approach optimality [5].

AMR control includes allocating tasks (optimal task assignment), localization (current location in map), path planning (collision free paths) and motion planning (obstacle avoidance) [6]. Of these control tasks, task allocation falls under the category of fleet management, while the rest fall under autonomous control. Here we mainly address the task allocation problem, while also considering its relation to path and motion planning, taking into account the AMR's localization and path planning capabilities.

Our approach for task allocation is partly centralized and partly distributed. A coordinator agent begins the task allocation as a negotiation process. The available AMRs then communicate their capability information to the coordinator, which attempts to maximize the system intelligence by maximizing the remaining entropy corresponding to the variety of tasks that the AMR fleet is capable of performing. This follows the idea of causal entropy, where intelligent behavior - hard to quantify as such - can be reduced to maximizing one's options [7].

Capability information, however, is based on the estimated performance of an AMR's individual task execution. For path following, higher accuracy results in minimizing the expected entropy. This is comparable to the approach of Valavanis, who proposes the use of entropy as a measure for an overall system, including the inter-level and intra-level (vertical and horizontal) measure of uncertainty and precision of task allocation and execution [8].

AMRs are currently well-suited for point-to-point navigation and transportation tasks. Here we consider extending the use of AMRs regarding their continuous path following capabilities, when precision is especially relevant, such as for cleaning along a desired path. The expected path execution accuracy is then linked to the probabilities of achieving the nominal paths. When viewed this way, the minimization of entropy translates into "how well" or "how precise" the algorithm or controller performs, and analogously, how precisely a task is executed [8].

The main novelty of our approach is the dualistic use of entropy for both multi-robot coordination and control of individual robots. The next sections introduce more details of this approach: entropy based task allocation (Section II), entropy based estimation of the task execution performance (Section III), and examples of different configurations of system and task entropies (Section IV). These are followed by the Discussion in Section V and Conclusion in Section VI.

## II. TASK ALLOCATION AND SYSTEM ENTROPY

### A. *Task allocation methods*

A variety of algorithms are proposed for task allocation. Optimization based algorithms search for an optimal solution, maximizing a profit or minimizing a cost using global information and considering all constraints. If the solution space is small, exact solutions may be found. Metaheuristic algorithms are partial search algorithms that may provide a sufficiently good solution, and are suited well when incomplete or imperfect information or limited computation capacity is available. Methods based on swarm intelligence are getting more and more popular, and one of the most promising ones is the Firefly Algorithm. It is a kind of stochastic, nature-inspired, meta-heuristic algorithm that can be applied for solving the hardest optimization problems (also NP-hard problems) [9, 10].

Market-based approaches use an auction, where each robot - an agent - uses its local information to set up and communicate a bid. Market-based approaches are robust and scalable and accept a lot of flexibility and complexity due to the distribution of computation. Because of their decentralized nature, robustness, and the possibility to scale to large and very complex systems is achieved without pushing the boundaries [6]. Based on experimental results metaheuristic algorithms have shown better scalability than market-based algorithms [11].

Discrete event simulation has also been applied in task allocation. In [12] the performance of an assembly line in the automotive industry has been analysed with centralized cloud-based systems, detecting material supply requirements, triggering material supply operations, scheduling them, and communicating schedules to the AMRs.

We are targeting flexibility and robustness in task allocation, and are applying market-based algorithms at the system level and simulations at the level of individual AMRs.

### B. *System specification*

The requested tasks are represented by a set $S$ of task labels for types of maintenance services

$$S = \{task_1, task_2, \ldots, task_n\} \tag{1}$$

The AMR system for maintenance services is composed of a set of AMR resources, i.e., set of agents $A$

$$A = \{agent_1, agent_2, \ldots, agent_m\} \tag{2}$$

Each agent $agent_i$ has an associated set of tasks it can carry out, i.e., capabilities $C_i$, such that $C_i \subseteq S$. Maintenance jobs are released to the maintenance service system as orders $O_k$ at time $k$, with each order specifying the type of task, the target for the service, and due time.

$$O_k = \{\text{“task_label”: } \langle task_k \rangle, \text{ ”target_label”: } \langle target_k \rangle, \text{ “due_time”:} \langle time_k \rangle \} \tag{3}$$

Targets are stored as lists according to their types. Targets are classified, for maintenance tasks, but here we consider only the class “*area*”:

$$T = \{ T_1, T_2, \ldots, T_N\}, \quad T_i = \{\text{”target_label”: } \langle target_k \rangle, \text{ ”target_type”: } \langle type \rangle \} \tag{4}$$

where $T$ is the set of targets, which is here a floor area, and such areas are defined by a list of bounding coordinates

$$\langle area_k \rangle = \{(x_1, y_1), (x_2, y_2), \ldots, (x_n, y_n)\} \tag{5}$$

which are assumed to compose a convex hull.

We apply a decentralized task allocation method in which AMRs negotiate and bid for task assignments. Coordination is carried out by a fleet manager agent, which receives task orders from a scheduler, with targets being areas on floors. The fleet manager uses a model (digital twin) to set up an auction, where all available (AMR) agents will participate. The agents use their own models (digital twins) about the environment (maps), and create bids with expected task execution entropy. In Fig. 1 the system structure is illustrated.

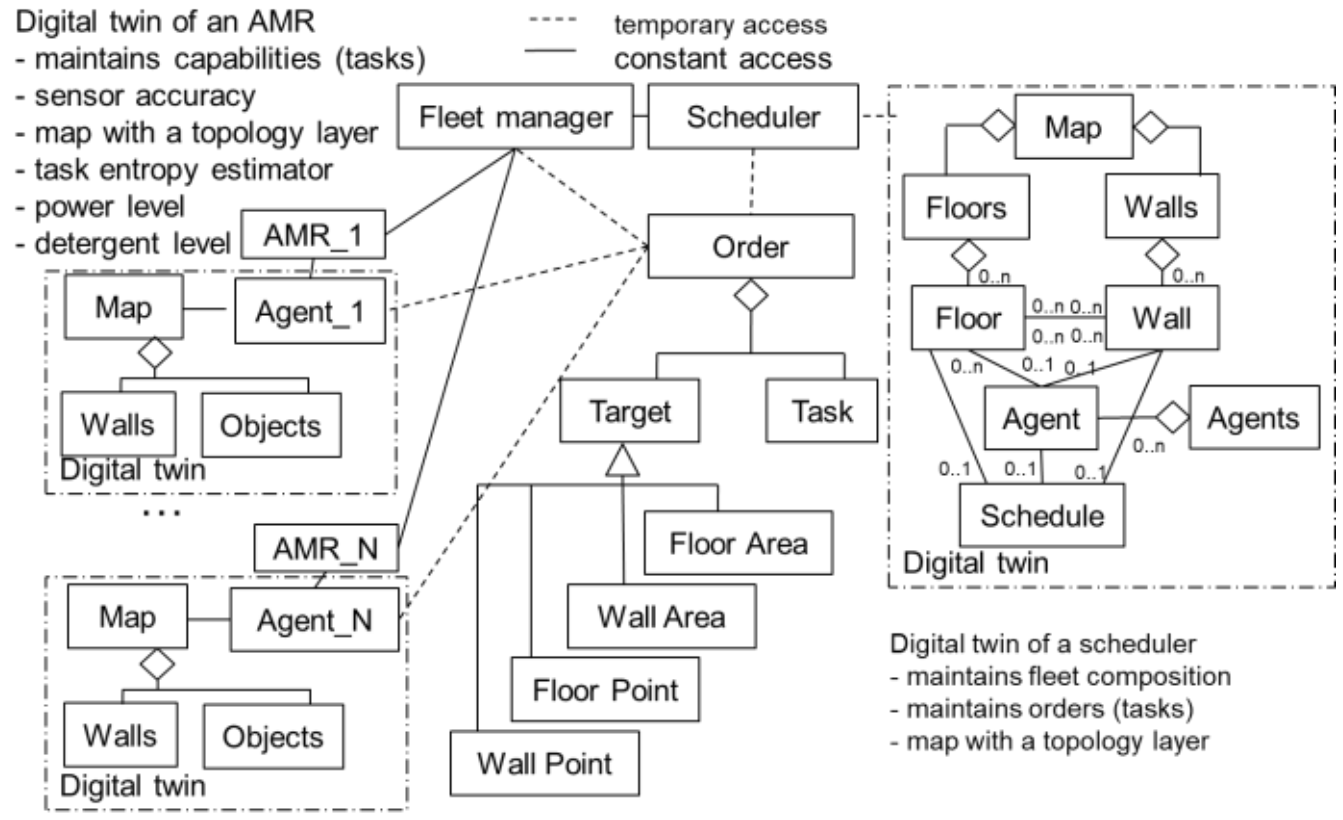

Figure 1. Multi-robot system model for maintenance tasks.

The ultimate goal of the task allocation is to maintain the capability of the system at the best level to respond to any type of requested tasks. This can be achieved best when after allocating an order to an agent, the capabilities of the set of remaining agents - considered as a system configuration - can cover as many tasks as possible. In this way the intelligence of the system (following the Causal Entropy Idea as specified by [7]) can be maintained at its maximum level. Mathematically this is done by maximizing the system entropy, i.e. the entropy of the system configuration after the task allocation. In case there are several candidate agents for task allocation, resulting in equal remaining system entropy, the agent with the best bid will be selected, i.e. the one that minimizes the expected task entropy.

Our task allocation algorithm is a revised version of the contract net protocol [13, 14]. In Fig. 2, our protocol implementation is illustrated. The Scheduler, in this case the Maintenance controller, sends a goal statement (task order) to the Fleet manager, which is responsible for the coordination of the task order. The Fleet manager starts an auction by announcing the task request to the Agents (AMRs). The Agents start their bidding process by requesting the task entropies for the task from their geometric Digital Twins (DT), based on their capability to perform the requested task. After getting the task entropy from the DT, the Agents send their capabilities to carry out tasks (list of task types) and bids for the requested task (entropy as a performance measure for the requested order) to the Fleet manager. Based on the entropies the Fleet manager awards the order to one agent. Awarding is based on candidate configurations without the awarded agent. The configuration with the highest system entropy is selected and the agent outside this configuration is awarded the order. In Section 3 the awarding process is presented in more detail.

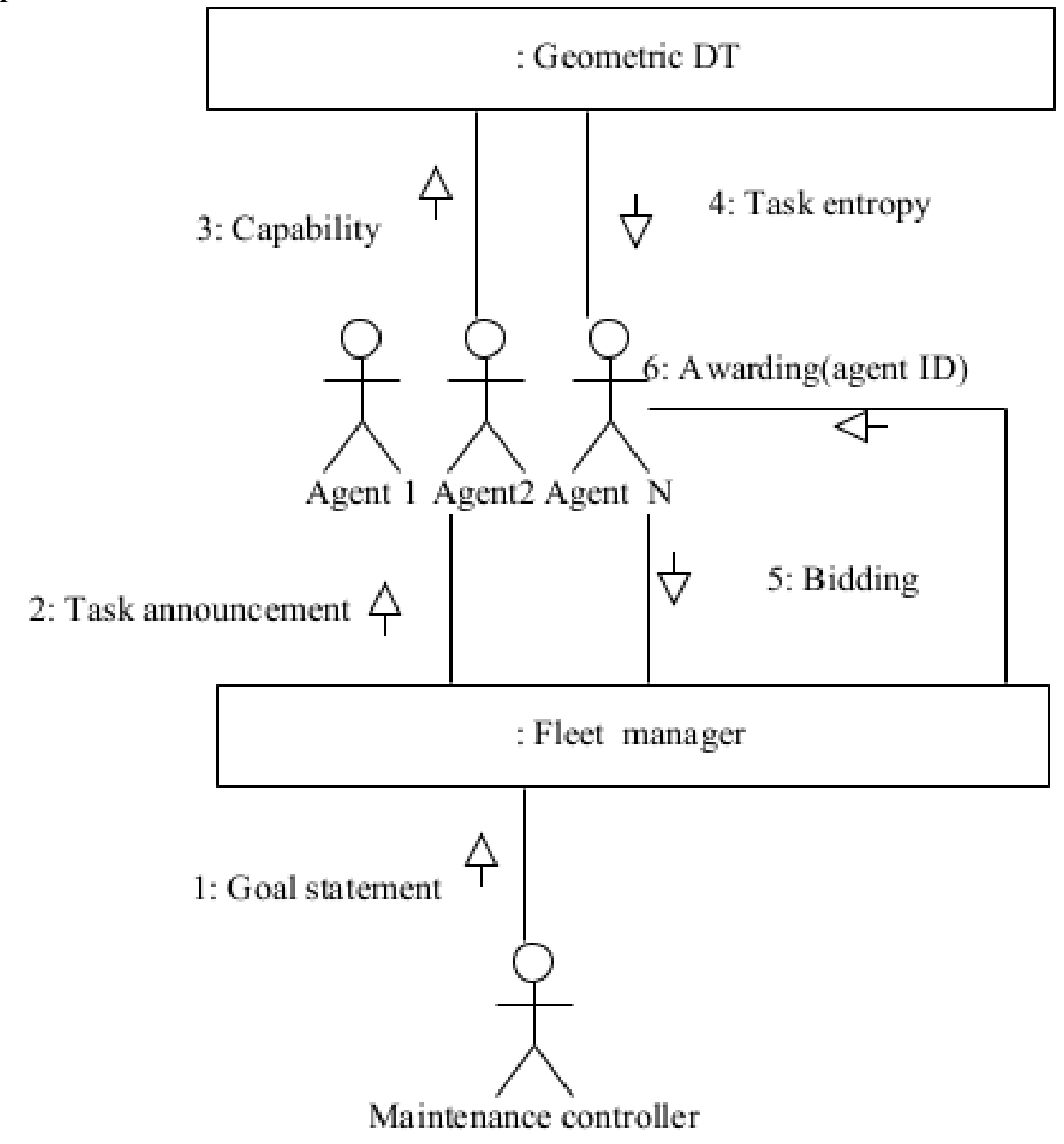

Figure 2. Negotiation protocol - Contract Net.

The goal for the maintenance service system is to be able to carry out an ordered job with minimum delay, i.e. to have an agent as often as possible available to carry out the requested task. For computing the system entropy, the probability with even distribution of having a capability to carry out the task $j$ is

$$P_j = n_{task_j} / n_{all_tasks} \tag{6}$$

where $n_{task_j}$ is the number of agents capable of carrying out task j

$n_{all_tasks}$ is the number of all tasks all the agents can carry out.

The system entropy, also reflecting the information contents of the set of capable tasks in bits, at time $t$ is defined then as:

$$H_t = -\sum_{j=1}^{n} P_{j,t} \log 2\, P_{j,t} \tag{7}$$

where $H_t$ is the system entropy considering the capability to carry out tasks at time $t$

$P_{j,t}$ is the probability for the system of having the capability to carry out task $j$ at time $t$

## III. TASK PLANNING AND TASK ENTROPY

### A. *Task performance: spatial uncertainty and entropy*

Our maintenance tasks rely on an AMR's ability to realize continuous paths based on visual navigation. Path point frames have position and orientation (=spatial) uncertainties in the floor segment frame. Although we model two translational and one rotational degrees of freedom (object in planar space), for simplicity, we count only traversal uncertainties in determining the task entropy.

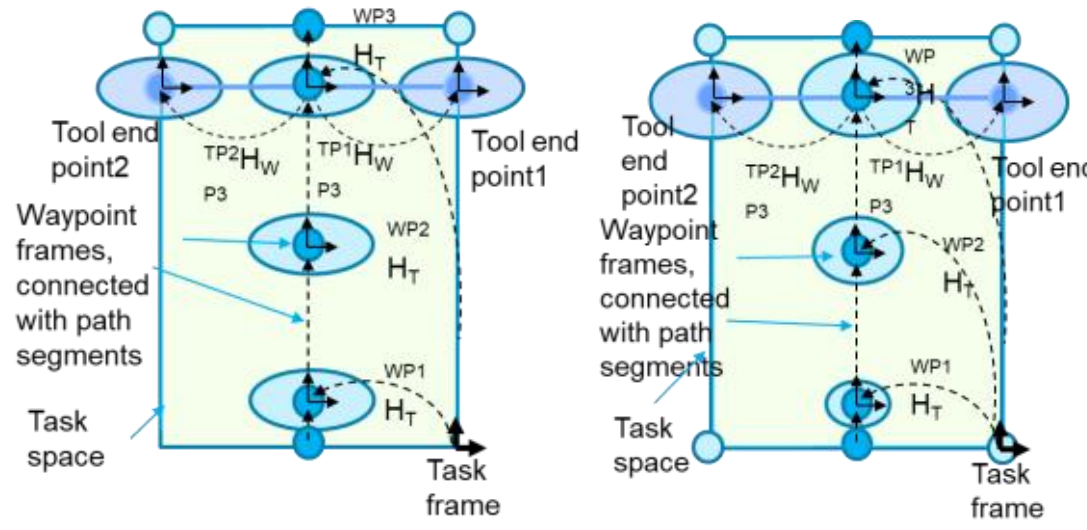

Figure 3. Location uncertainties of tool endpoints and the robot body frame for visual navigation mode (left) - the level of uncertainty remains approximately equal for all the path points, and for odometry navigation (right) - the level of uncertainty gradually increases for the path points.

We estimate the AMR's path execution accuracy for each path point, from which we get probabilities, how well the realized paths can cover target areas. Task uncertainties are modelled as uncertainties of tool endpoints in path points (see Fig. 3) This reflects to the need of having overlapping path segments to reliably cover the target area: a lower uncertainty level leads to less need for the overlapping of paths. Probabilities of tool point locations will lead us to estimate the entropy in realizing the task.

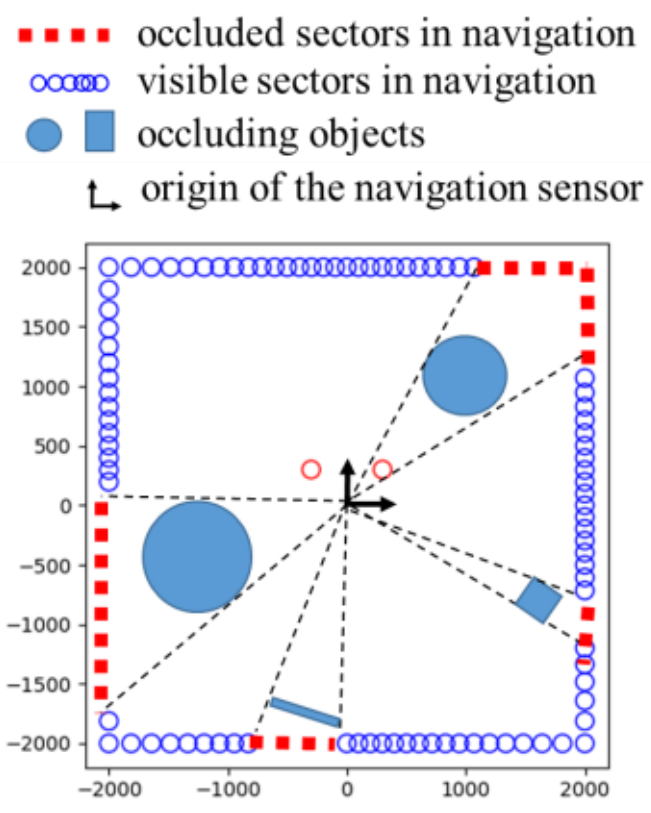

Figure 4. Navigation scene example: nominal location of tool points (red circles in the center), origin of the navigation sensor of the AMR (black coordinate frame), visible sectors of the navigation map (blue circles) and occluded / non-visible parts of the navigation map (red square dots).

The agents are using their digital twins, with a navigation map for creating bids for a task in an order. The navigation map is an on-board local copy from the environment map at some point of time. The agents plan the tasks as paths covering the requested areas and compare their local navigation maps to a common map. The digital twin is used to keep track of the dynamic situation on the shop floor. If new objects have appeared into the environment after the local map has been created, then they are blocking the visibility to the reference points represented in the local navigation maps, thus reducing the amount of reference data for localization in navigation. This is illustrated in Fig. 4.

For the maintenance tasks (e.g., vacuum cleaning, wiping, detergent spreading) we assume the tool to be attached to the front part of the AMR and to be longitudinal. The locations of the tool endpoints on the floor along nominal path points determine the tool coverage on the floor. For simplicity, we assume the width of the area to be same as the tool length. Because of location uncertainties in navigation, one end of the tools gets outside the nominal area; and this uncertainty is roughly at the same level through all points while visually navigating (Fig. 3, left), and increases gradually if navigating with odometry data (Fig. 3, right). Estimating the expected amount of the “missing” part of the tool from the task area, as tool endpoint uncertainties, gives us the probabilities to cover the nominal area in the transverse direction. The entropy, calculated from the coverage probability, gives a measure how much the real tool endpoint deviates from the nominal tool endpoint: the smaller the entropy, the closer the real point to the nominal point.

Uncertainties of the tool point locations are modelled as Gaussian distributions of the parameters, as the 1st std deviation $\sigma_{1st}$ in the transverse direction. This can be represented in the global floor segment frames, or conditionally in the earlier waypoint of the path of the task (odometry based navigation).

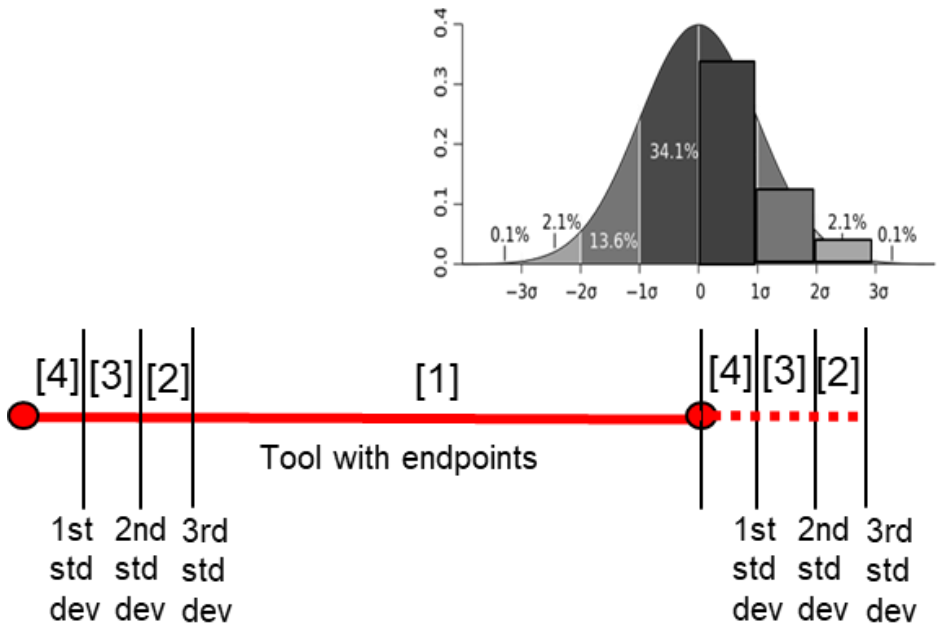

Figure 5. Location uncertainty tool coverage with location std dev σ: 50 % of samples [1] inside the nominal location, 34.1 % of samples [4] outside from endpoint to endpoint + σ, 13.6 % of samples [3] outside from endpoint + σ to endpoint + 2σ, 2.1 % of samples [2] outside from endpoint + 2σ to endpoint + 3σ.

All the variations in realized path points, compared to nominal path points, causes part of the tool to miss the target. How much the part of the line between the tool endpoints hits the area of nominal tool endpoint locations can be analysed using the standard deviations (see Fig. 5), here considering the transverse direction. For the tool endpoint location, there is a 50 % probability to be located in the range, scaled to the length of the tool (area [1] in Fig. 5):

$$R1 = [0, l_{tool} - 3\sigma] / l_{tool} \quad (8)$$

As a probability for a point to be located in this range, we get

$$P1 = |R1| \quad (9)$$

For the endpoint location, there is 2.1 % probability to be located in the range, scaled to the length of the tool (area [2] in figure 5)

$$R2 = [l_{tool} + 2\sigma, l_{tool} + 3\sigma] / l_{tool} \quad (10)$$

As a probability for a point to be located in this range, we get

$$P2 = 0.021\,|R2| \quad (11)$$

For the endpoint location, there is 13.6 % probability to be located in the range, scaled to the length of the tool (area [3] in Fig. 5):

$$R3 = [l_{tool} + \sigma, l_{tool} + 2\sigma] / l_{tool} \quad (12)$$

As a probability for a point to be located in this range, we get

$$P3 = 0.136\,|R3| \quad (13)$$

For the endpoint location, there is 34.1 % probability to be located in the range, scaled to the length of the tool (area [4] in Fig. 5)

$$R4 = [l_{tool}, l_{tool} + \sigma] / l_{tool} \quad (14)$$

As a probability for a point to be located in this range, we get

$$P4 = 0.341\,|R4| \quad (15)$$

The entropy for accomplishing the task concerning spatial uncertainty within one path point comes then as:

$$H_{task} = -(P1 \log_2(P1) + P2 \log_2(P2) + P3 \log_2(P3) + P4 \log_2(P4)) \quad (16)$$

For a complete task with N path points, we consider the sum of path point entropies, as the upper bounds of total task entropy [14]:

$$H_{wp1,\ldots,wpN} = \max\{H_{wp1}, \ldots, H_{wpN}\} \leq H_{wp1} + \ldots + H_{wpN} \quad (17)$$

where $H_{wp1,\ldots,wpN}$ is the upper bound of the total task entropy

$H_{wpi}$ is the entropy for path point i, i = 1 to N.

### B. Estimation of the navigation uncertainty

For the entropy of a task, as given above, we need an estimate about the spatial uncertainty as a normal distribution. We simulate the uncertainty of location estimation for LIDAR based navigation, when all or a subset of navigation points (of interest in the map) are visible and known. We match visible modelled points, extracted from the map, to simulated visible points as they would be measured from the environment. Each measured point has the following relation with the corresponding point in the map:

$$P_{i,\,measured} = T + R\,P_{i,\,nominal} \quad (18)$$

where $T$ is the translation vector and

$R$ is a 3x3 rotation matrix, describing the rotation between the base frame and the measured frame.

The translation between the measured and the base frame is the mean of the measured points:

$$P_{mean,nominal} = \frac{1}{n}\sum_{i=1}^{n} P_{i,nominal} \quad (19)$$

$$P_{mean,measured} = \frac{1}{n}\sum_{i=1}^{n} P_{i,measured} \quad (20)$$

The covariance between the nominal and measured points is then

$$C = (P_{nomial} - P_{mean,nominal})(P_{measured} - P_{mean\ measured})^T \quad (21)$$

The rotation matrix can then be acquired using singular value decomposition (for more details, see [15]):

$$U,S,V = SVD(C) \quad (22)$$

where $U$, and $V$ are orthogonal matrices and S is a unitary matrix.

The corrected rotation matrix can be determined as

$$R_{corr} = V\,U^T \quad (23)$$

The translation can then be derived

$$T = P_{mean\ measured} - R_{corr}\,P_{mean,nominal} \quad (24)$$

We simulate the AMR's navigation as estimating the pose of the AMR in each path point of the task, using equations (19) - (24) several times, with added Gaussian noise in the simulated measured range values. The standard deviation (variance/covariance) for translation parameters are determined for entropy calculations. Some additive uncertainty models are available, e.g., accumulation with Kalman filters [16], which should be considered if other types of navigation, like odometry, were applied.

## IV. Simulation tests

### A. Test system

We tested our approach with simulation tests. Fig. 6 shows the SW architecture of our test system. The system has four main classes: Agent (Python class), Fleet manager (Python class), geometric DT (Coppelia robot simulator platform) and discrete event simulator (Simpy – discrete event simulator framework in Python [17]). A separate scheduler and the related Digital Twin are to be implemented in the future. Some of the Agents' and Fleet manager's methods are run as Simpy processes.

The Agent has attributes for a mobile robot's Id, name, capabilities (maintenance tasks it can carry out), Simpy environment reference and task entropies. The Agent has methods for bidding, capability (for requesting task entropies from the DT) and a method for showing the status information of the agent. The last one is running as a Simpy environment process. The Fleet manager has attributes for an Id, a list of each agent's and the Simpy environment reference. The Fleet manger has methods for requesting bids from agents, awarding, and showing status information. In the current version, the scheduler method is used for coordinating the task order process. The scheduler and status methods are running as Simpy environment processes.

The Simpy discrete event simulator has two methods, one for setting up the Simpy environment and one for running the simulation. The Geometric simulator is implemented with CoppeliaSim [18], and has an interface for controlling the model objects from external Python SW.

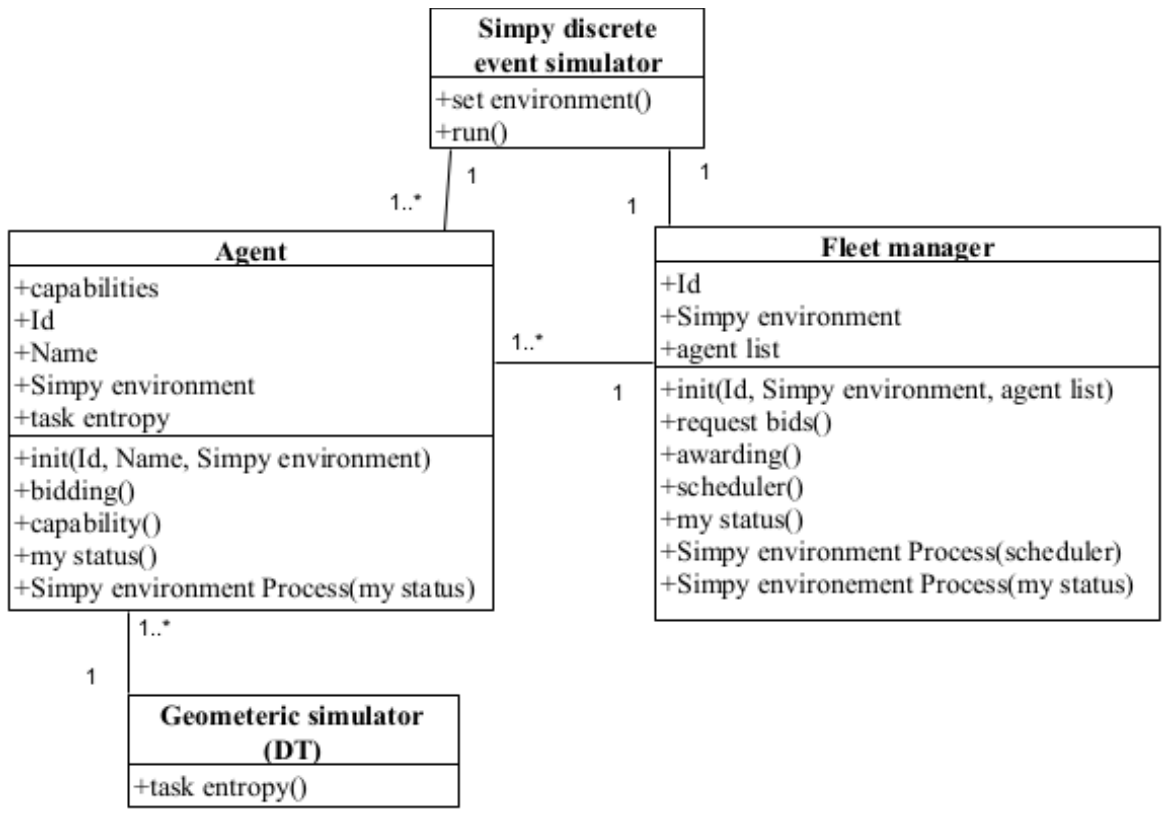

Figure 6. Main architecture of the implemented simulation system.

### B. System entropy

The test system was composed of five simulated mobile robots, carrying out five maintenance tasks:

$S$ = *{garbage_collect, vacuum_clean, mop_floor, wax_floor, disinfect_floor}* (25)

where $S$ is the set of maintenance task labels.

Examples for different system configurations were considered, and exemplary system entropies were calculated using equations (6) and (7). Three configurations with corresponding entropies are given in Fig. 7. Maximum entropy is attained when the number of tasks each agent can do is equal for all agents, and task coverage even (2.32 bits).

| | garbage_collect | vacuum_clean | mop_floor | wax_floor | disinfect_floor |
|---|---|---|---|---|---|
| robot1 | 1 | | | | 1 |
| robot2 | 1 | 1 | | | |
| robot3 | | 1 | 1 | | |
| robot4 | | | 1 | 1 | |
| robot5 | | | | 1 | 1 |

**H=2.321928 bits**

| | garbage_collect | vacuum_clean | mop_floor | wax_floor | disinfect_floor |
|---|---|---|---|---|---|
| robot1 | 1 | | | | |
| robot2 | 1 | 1 | | | |
| robot3 | 1 | 1 | 1 | | |
| robot4 | 1 | 1 | 1 | 1 | |
| robot5 | 1 | 1 | 1 | 1 | 1 |

**H=2.149256 bits**

| | garbage_collect | vacuum_clean | mop_floor | wax_floor | disinfect_floor |
|---|---|---|---|---|---|
| robot1 | 1 | 1 | | | |
| robot2 | 1 | 1 | | | |
| robot3 | 1 | 1 | | | |
| robot4 | 1 | 1 | | | |
| robot5 | 1 | 1 | 1 | 1 | 1 |

**H=1.914341 bits**

Figure 7. Multi-robot system configurations: when all agents can do an equal number of tasks, the entropy is at its maximum: 2.3219 bits.

### C. Task entropy

Simulations were made to test the effect of navigation on task entropy. This included the effect of additional obstacles, which occlude the available navigation wall points, and also the effect of navigation point range noise. In general, it is expected that worse navigation/positioning would result in higher task entropy.

A simple task was simulated consisting of an AMR driving along a straight path with 5 path points, which is analogous to part of a floor cleaning task (see Fig. 8). The simulations were made in the CoppeliaSim robot simulation environment, linked with external Python scripts controlling the simulation. The AMR has a top mounted 1D laser scanner which can rotate 360° to make range measurements to build a 2D map, and to navigate. The simulated test environment is a square room with 3 m-long walls, also containing various cylindrical obstacles. It is assumed that the robot has an initial map of the room with no obstacles, however during the task, the obstacles occlude parts of the walls.

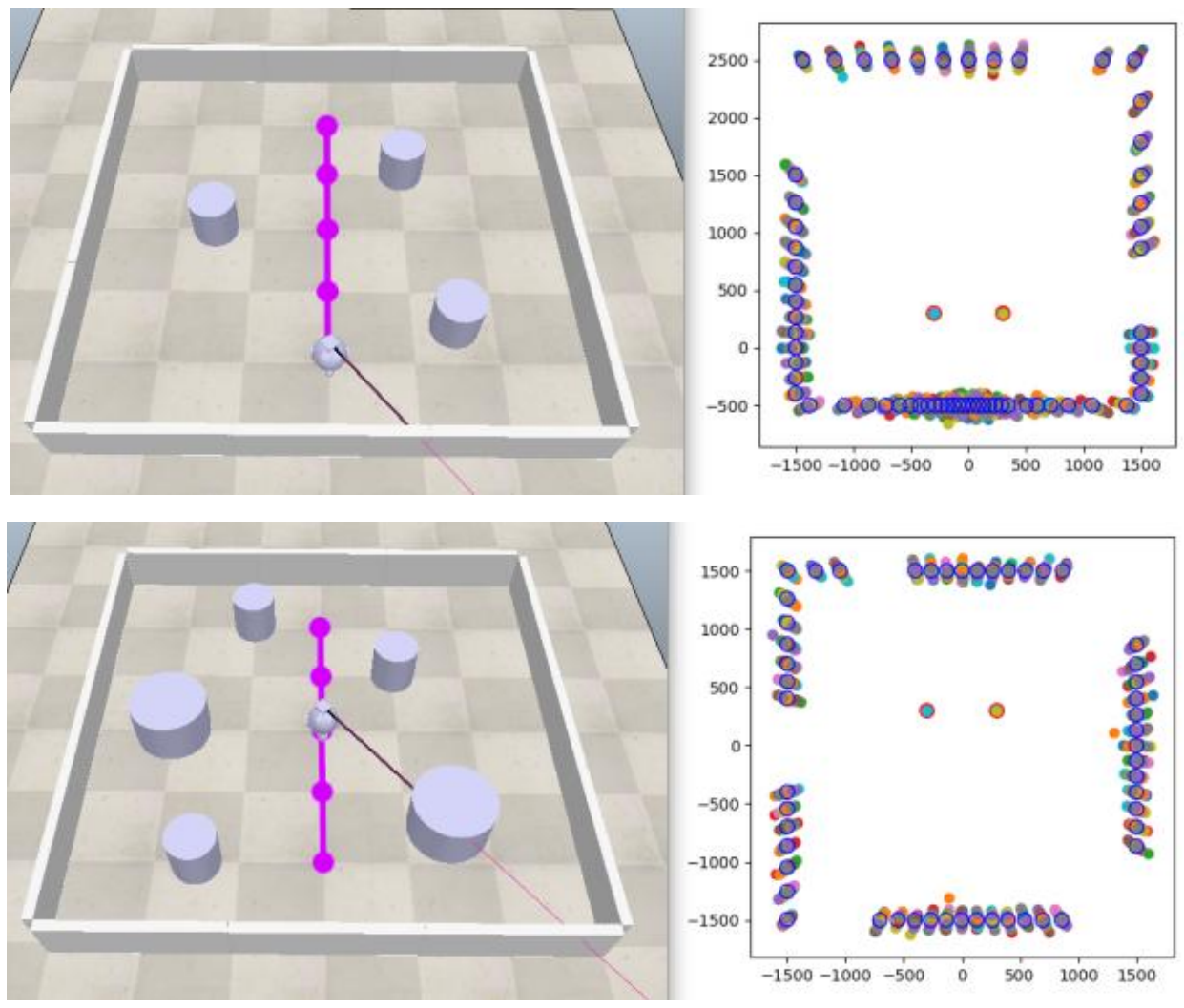

Figure 8. Mobile robot driving along straight path with 5 waypoints in Scenario 1 (top) and Scenario 2 (bottom). A new laser scan is made at each waypoint (shown at right for current waypoint).

During the task, the robot begins at the bottom end of the path (Fig. 8 top), then is translated forward by one 0.5 m increment to the next path point. At each point, a new scan is made of the room by rotating the laser with 5° increments. Examples of the maps produced are shown at right in Fig. 8. The blue circles in the maps represent the nominal wall points, while the coloured points represent a Gaussian range error which was introduced, initially with a standard deviation of 50 mm. Location estimation using equations (20) - (25) was carried out, with random noise added to the range values in each estimation cycle We assume point-to-point matches between nominal and measured points, though this is finally not practically so because of locating errors. Instead, further processing of points and local features like corners should be done, to get a set of interest points for reliable point matching. However, at this stage we are content with our simplification as it results in the desired measure of uncertainty and reflects the reality. The localizing test was repeated 20 times, which is the reason for the spread of points at each nominal wall point. The occlusion from the obstacles is evident from the gaps in the mapping wall points.

To test how the amount of mapping information affects the task entropy, two scenarios were created of the same task in the same room, but with different surrounding obstacles. Fig. 8 (top) shows Scenario 1, with three cylindrical obstacles in the room, each 0.33 m diameter. In this figure the robot is at the initial path point. Fig. 8 (bottom) shows Scenario 2, with five obstacles, two of them with 0.5 m diameter and three with 0.33 m diameter. In this figure the robot is at the 3$^{rd}$ path point.

As the robot moved along the line, the task entropy was calculated at each of the 5 path points using Equation 16, using the tool point position error standard deviation. The entropy values at each point were then summed up to get a total task entropy. Table 1 summarizes the results for the two different scenarios. As expected, the task entropy was higher for Scenario 2 which had fewer mapping points available due to more occlusion from the obstacles.

Table 1: Total Task Entropy for Scenario 1 and Scenario 2

| | Scenario 1 | Scenario 2 |
|---|---|---|
| Total Task Entropy [bits] | 0.3011 | 0.3413 |

To test the effect of the range noise on the task entropy, the task was simulated in Scenario 1 with the range noise standard deviation reduced from 50 mm (used previously) to 30 mm and 10 mm. Fig. 9 shows two maps from the 3$^{rd}$ path point, with a range noise of 50 mm (left) and 10 mm (right).

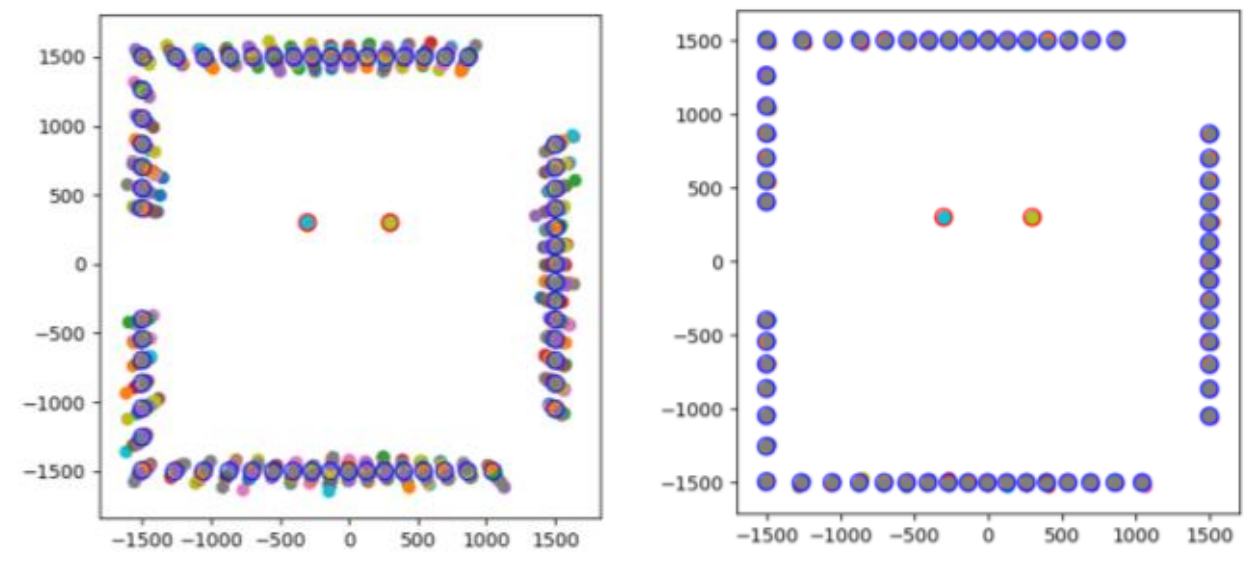

Figure 9. Map created at Point 3 during Scenario 1 with range noise of 50 mm (left) and 10 mm (right).

Table 2 summarizes the results with the different range noises, using the same method as before to calculate the total task entropy. As expected, this value decreases with decreasing range noise, since noisier measurements result in higher positioning error.

Table 2: Total Task Entropy for Scenario 1 with Proximity Sensor Range Noise Variation

| | Sensor Range Noise (mm) | | |
|---|---|---|---|
| | 50 | 30 | 10 |
| Total Task Entropy [bits] | 0.2910 | 0.2060 | 0.0816 |

## V. DISCUSSION

We have used entropy as a criterion to coordinate a multi-robot system, having priority in flexibility rather than optimal performance. From the system point of view, as wide as possible coverage of tasks is maintained by maximizing system entropy. We apply a negotiation protocol for this, and this scales up well because of its distributed nature. This shares ideas of system control of self-organizing sensors, where the energy problem has been addressed by reducing the difference in remaining energy among nodes [19]. Interestingly enough, system capabilities can also be found already even in existing systems engineering standardization, as ISA-95 has the “capability” attribute defined (the “OperationsCapability” structure, for more details see e.g. [20]).

Performance for individual task orders is considered by minimizing the total task entropy. Expected path execution accuracy finally sets the performance level for an AMR - and further the entropy - and tells how much of the required area can be expected to be covered by the tool and reflects indirectly the need to overlap the paths for reliable coverage of the requested area. This compares to the ideas of situated entropy with digital twins, sensors and actuators [21]. We see path following accuracy as an important aspect when considering the wider use of current industrial AMRs, for which tasks with continuous paths have had so far a minor role.

## VI. Conclusion

Entropy can be used in two ways as a performance measure in the coordination of operations of an AMR system. We presented a novel approach for task scheduling and assignment, and gave preliminary results based on simulations showing the feasibility of our approach. In the future we will go for more complicated system setups and scenarios, and extend our system with a digital twin for the scheduler.

## AcknowledgmentS

The authors would like to thank Dr. Stephen Fox of VTT and Prof. Juha Plosila and Dr. Hashem Haghbayan from the University of Turku for discussions of entropy and simulation methodologies.